\documentclass{article} 
\usepackage{hyperref,psfrag}
\usepackage{url}
\usepackage{url,cite}
\usepackage[outdir=./]{epstopdf}
\usepackage{graphicx,float,pgfplots,wrapfig,sidecap,lipsum}
\usepackage{tabularx}
\usepackage{booktabs}
\usepackage{paralist}
\usepackage{algorithm}
\usepackage[noend]{algorithmic}
\usepackage[pass]{geometry}
\usepackage{amsfonts,amsthm,amsmath,amssymb}
\usepackage{enumitem}
\usepackage{xcolor}
\usepackage[font=normal,labelfont=bf]{caption}
\usepackage{mathtools} 

\usepackage{tablefootnote}
\usepackage[caption=false]{subfig}

\usepackage{tikz}
\usetikzlibrary{fit}
\usetikzlibrary{calc,shapes}
\usetikzlibrary{decorations.pathmorphing} 
\usetikzlibrary{fit}					
\usetikzlibrary{backgrounds}	

\def\Rbb{\mathbb{R}}
\def\Ebb{\mathbb{E}}
\def\tha{{\mbox{\tiny th}}}

\newcommand{\ouralgorithm}{\mathsf{CT}}

\newcommand{\bp}{\begin{psfrags}}
\newcommand{\ep}{\end{psfrags}}
\newcommand{\bc}{\begin{center}}
\newcommand{\ec}{\end{center}}

\newcommand{\Conv}{\mathop{\scalebox{1.5}{\raisebox{-0.2ex}{$\ast$}}}}%
\newcommand{\Toep}{\mathsf{Cir}}
\newcommand{\CToep}{{\mathcal{F}}}
\newcommand{\CCir}{{\mathcal{F}}}
\newcommand{\Diag}{\mathsf{diag}}

\newcommand{\DFT}{\mathsf{FFT}}
\def\bfU{{\mathbf U}}

\DeclareMathOperator{\flatten}{unfold}
\DeclareMathOperator{\modeA}{\mathcal{F}}
\DeclareMathOperator{\modeB}{\mathcal{G}}
\DeclareMathOperator{\modeC}{\mathcal{H}}

\DeclareMathOperator{\Cum}{C_3}

\def\bfI{{\mathbf I}}
\DeclareMathOperator{\newCum}{M}
\def\nn{\nonumber}

\DeclareMathOperator{\Fbb}{\mathbb{F}}
\DeclareMathOperator{\opt}{opt}
\DeclareMathOperator{\block}{blk}
\DeclareMathOperator{\blkdiag}{Blkdiag}

\newtheorem{theorem}{Theorem}[section]
\newtheorem{lemma}[theorem]{Lemma}

\floatname{algorithm}{Procedure}

\def\beq{\begin{equation}}

\def\eeq{\end{equation}\noindent}
\newcommand*{\hermconj}{^{\mathsf{H}}}

\usepackage{fullpage}

\title{Convolutional Dictionary Learning through Tensor Factorization}
\date{}
\author{Furong Huang \thanks{University of California Irvine, Department of Electrical Engineering and Computer Science, furongh@uci.edu} \and
Animashree Anandkumar \thanks{University of California Irvine, Department of Electrical Engineering and Computer Science, a.anandkumar@uci.edu}}

\begin{document}
\maketitle
\begin{abstract}Tensor methods have emerged as a powerful paradigm for consistent learning of many latent variable models such as topic models,  independent component analysis and dictionary learning. Model parameters are estimated via CP decomposition of the observed higher order input moments. However, in many domains, additional invariances such as shift invariances exist, enforced via models such as convolutional dictionary learning. In this paper, we  develop novel tensor decomposition algorithms for parameter estimation of convolutional models. Our algorithm is based on the popular alternating least squares method, but with efficient projections onto the space of stacked circulant matrices. Our method is embarrassingly parallel and consists of simple operations such as fast Fourier transforms and matrix multiplications. 
Our algorithm converges to the dictionary much faster and more accurately compared to the alternating minimization over filters and activation maps.
\end{abstract}
\section{Introduction}
The convolutional dictionary learning model posits that the input signal  $x$ is generated as   a linear combination of convolutions of unknown dictionary elements or {\em  filters} $f_1^*, \ldots f_L^*$ and unknown  {\em activation maps} $w_1^*, \ldots w_L^*$:
\beq\label{eqn:sparsedef} x = \sum\limits_{i\in [L]}  f_i^* \Conv w_i^*,\eeq where $[L]:=1,\ldots, L$. The vector $w_i^*$ denotes the activations at locations, where the corresponding filter $f_i^*$ is active.  Convolutional models are ubiquitous in machine learning for image, speech and sentence representations~\cite{zeiler2010deconvolutional,kavukcuoglu2010learning,bristow2013fast},
and in neuroscience for modeling neural spike trains~\cite{olshausen2002sparse,ekanadham2011blind}. Deep convolutional neural networks are a multi-layer extension of these models with non-linear activations. Such models have revolutionized performance in image, speech and natural language processing~\cite{zeiler2010deconvolutional,kalchbrenner2014convolutional,kalchbrenner2014convolutional}. 

In order to learn the model in \eqref{eqn:sparsedef}, usually a square loss reconstruction criterion is employed: \beq\label{eqn:alt-min}\min_{f_i,w_i: \|f_i\|=1} \|x - \sum\limits_{i\in [L]}  f_i \Conv w_i\|^2 .\eeq The constraints $(\|f_i\|=1)$  are enforced, since otherwise, the scaling can be exchanged between the filters $f_i$ and the activation maps $w_i$. Also,  an additional regularization term is usually added to the above objective, e.g.  to promote sparsity on $w_i$, an $\ell_1$ term is added.

A popular heuristic for solving \eqref{eqn:alt-min}   is based on  alternating minimization~\cite{bristow2014optimization}, where the filters $f_i$ are optimized, while keeping the activations $w_i$ fixed, and vice versa.  Each alternating update can be solved efficiently (since it is linear in each of the variables). However,  the method is  computationally   expensive in the large sample setting since  each iteration requires a pass over all the samples, and in modern machine learning applications, the number of samples can run into billions.
Moreover, the alternating minimization has multiple local optima and reading the   global optimum of~\eqref{eqn:alt-min} is NP-hard in general. This problem is severely amplified in the convolutional setting due to additional symmetries, compared to the usual dictionary learning setting (without the convolutional operation). Due to shift invariance of the convolutional operator, shifting a filter $f_i$ by some amount, and applying a corresponding negative shift on the activation $w_i$ leaves the objective in \eqref{eqn:alt-min} unchanged. Thus, solving \eqref{eqn:alt-min} is fundamentally   {\em ill-posed} and has a large number of equivalent solutions.  On the other hand, imposing shift-invariance constraints directly on the objective function in \eqref{eqn:alt-min} results in non-smooth optimization which is challenging to solve.

Can we design methods that  efficiently incorporate the shift invariance constraints into the learning problem?    Can they {\em break the symmetry} between equivalent solutions?  Are  they scalable to huge datasets? In this paper, we provide positive answers to these questions.
We propose a novel framework for  learning convolutional models through tensor decomposition.  We consider inverse method of moments to estimate the model parameters via decomposition of higher order (third or fourth order) input cumulant tensors. When the inputs $x$ are generated from a convolutional model in \eqref{eqn:sparsedef}, with independent activation maps $w_i^*$, i.e. a convolutional ICA model, we show that the moment tensors have a CP decomposition, whose components form a stacked {\em circulant} matrix. 
We propose a novel method for tensor decomposition when such circulant (i.e. shift invariance) constraints  are imposed.

Our method is a constrained form of the popular alternating least squares (ALS) method for tensor decomposition\footnote{The ALS method for tensor decomposition is not to be confused with the alternating minimization method for solving \eqref{eqn:alt-min}. While \eqref{eqn:alt-min} acts on data samples, ALS operates on averaged moment tensors.}. We show that the resulting optimization problem (in each ALS step) can be solved in closed form, and uses simple operations such as Fast Fourier transforms (FFT) and matrix multiplications. These operations have a high degree of parallelism: for estimating $L$ filters, each of length $n$, we require $O(\log n +\log L)$ time and $O(L^2n^3)$ processors. Thus, our method is embarrassingly parallel and scalable to huge datasets. We carefully optimize computation and memory costs by exploiting tensor algebra and circulant structure. We implicitly carry out many of the operations and do not form large (circulant) matrices to minimize storage requirements.


Moreover, our method requires only one pass over data to compute the higher order cumulant of  the input data or its approximation through sketching algorithms.  This is a huge saving in running time compared to the alternating minimization method in \eqref{eqn:alt-min} which requires a pass over data in each step. Decoding all the activation maps $w_i$ in each step of \eqref{eqn:alt-min} is hugely expensive, and our method avoids it by estimating only the  filters $f_i$ in the learning step. In other words, the activation maps $w_i$'s are averaged out in the input cumulant. After filter estimation, the activation maps are easily estimated using \eqref{eqn:alt-min} in one data pass.

Experiments further demonstrate superiority of our method compared to alternating minimization. Our algorithm converges accurately and much faster to the true underlying filters compared to alternating minimization. Our algorithm is also orders of magnitude faster than the alternating minimization. 




\subsection{Related Works}
The special case of \eqref{eqn:sparsedef} with one filter $(L=1)$ is a well studied problem, and is referred to as {\em blind deconvolution}~\cite{hyvarinen2004independent}.  In general, this problem is not identifiable, i.e. multiple equivalent solutions can exist~\cite{choudhary2014sparse}. It has been documented that in many cases alternating minimization produces trivial solutions, where the filter $f=x$ is the signal itself and the activation is the   identity function~\cite{levin2009understanding}. Therefore, alternative techniques have been proposed, such as
convex programs, based on nuclear norm minimization~\cite{ahmed2014blind} and  imposing hierarchical Bayesian priors for activation maps~\cite{wipf2013revisiting}. 
However, there is no analysis for settings with more than one filter.  Incorporating Bayesian priors has shown to reduce the number of local optima, but not completely eliminate them~\cite{wipf2013revisiting,krishnan2013blind}. Moreover, Bayesian techniques are in general more expensive than  alternating minimization in~\eqref{eqn:alt-min}.

The extension of blind deconvolution to multiple filters is known as convolutive blind source separation or convolutive independent component analysis (ICA)~\cite{hyvarinen2004independent}. Previous methods directly reformulate convolutive ICA as an ICA model, without incorporating the shift constraints. Moreover, reformulation leads to an increase of number of hidden sources from $L$ to $nL$ in the new model, where $n$ is the input dimension, which are harder to separate and computationally more expensive. 
Complicated interpolation methods~\cite{hyvarinen2004independent} overcome these indeterminacies. In contrast, our method avoids all these issues. We do not perform Fourier transform on the input, instead, we employ FFTs at different iterations of our method to estimate the filters efficiently.

The dictionary learning problem without convolution has received much attention. Recent results show that simple iterative methods can learn the globally optimal solution~\cite{AnandkumarEtal:COLT14,arora2013new}. In addition, tensor decomposition methods provably learn the model, when the activations are independently drawn (the ICA model)~\cite{anandkumar2014tensor} or are sparse (the sparse coding model)~\cite{anandkumar2014provable}. 
In this work, we extend the tensor decomposition methods to efficiently incorporate the shift invariance constraints imposed by the convolution operator.







\section{Model and Formulation}\label{sec:model}
\paragraph{Notation: }Let $[n]:=\{1,2,\ldots,n\}$. For a vector $v$,   denote the $i^{\tha}$ element as $v(i)$. For a matrix $M$,   denote the $i^{\tha}$  row as $M^i$ and   $j^{\tha}$  column as $M_j$.  For a tensor   $T\in\mathbb{R}^{n\times n\times n}$, its $(i_1,i_2,i_3)^{\tha}$ entry is denoted by $[T]_{i_1,i_2,i_3}$.
A \emph{column-stacked} matrix $M$ consisting of $M_i'$s  (with same number of rows) is $M := [M_1,M_2,\ldots,M_L]$. Similarly, a \emph{row-stacked} matrix $M$ from $M_i'$s (with same number of columns)  is $M:= [M_1;M_2;\ldots;M_L]$.

\paragraph{Cyclic Convolution: }The 1-dimensional (1-D) $n$-cyclic convolution $f\Conv w$ between vectors $f$ and $w$ is defined as $ \label{eqn:cyclic}v=f\Conv_n w, \ v(i) =\sum_{j\in [n]} f(j) w((i-j+1)\mod n).$ On the other hand, linear convolution is the combination without the modulo operation (i.e. cyclic shifts) above.  $n$-Cyclic convolution is equivalent to linear convolution, when $n$ is at least twice the support length of both $f$ and $w$~\cite{oppenheim1997signals}, which will be assumed. We drop the notation $n$ in $\Conv$ for convenience.
%
Cyclic convolution in \eqref{eqn:cyclic} is equivalent to $f\Conv w = \Toep(f)\cdot w, $
and \begin{equation}\label{eq:circulant}
\Toep(f) : = \sum_{p}f(p) G_p\in \Rbb^{n\times n} ,\quad  \left(G_p\right)^i_j : = \delta\left\{\left((i-j)\mod n\right)=p-1\right\}, \quad \forall p\in[n].
\end{equation} defines a  circulant matrix.   A circulant matrix $\Toep(f)$   is characterized by the vector $f$, and each column corresponds to a cyclic shift of $f$.
\paragraph{Properties of circulant matrices: }
Let $F$ be  the discrete Fourier transform matrix  whose $(m,k)$-th entry is $F_{k}^m = \omega_n^{(m-1)(k-1)}$, $\forall m,k\in[n]$ where $\omega_n = \exp(-\frac{2\pi i}{n})$. 
If $U : = \sqrt{n} F^{-1}$, $U$ is the set of eigenvectors for all $n \times n$  circulant matrices~\cite{gray2005toeplitz}. Let the  Discrete Fourier Transform  of a vector $f$ be $\DFT(f)$,  we express  the circulant matrix $\Toep(f)$ as  
 \begin{equation}\label{eq:circulant_fft}
 \Toep(f) = U \Diag(F\cdot f) U\hermconj = U \Diag(\DFT(f)) U\hermconj.
 \end{equation}   This is an important property we use in algorithm optimization to improve computational efficiency.

\paragraph{Column stacked circulant matrices: }We will extensively use column stacked circulant matrices $\mathcal{F}:=[\Toep(f_1),\ldots, \Toep(f_L)]$, where $\Toep(f_j)$ is the circulant matrix corresponding to filter $f_j$. 

\subsection{Convolutional Dictionary Learning/ICA Model}\label{subsec:ConvolutionalICAmodel}
%
%

\begin{figure}[!htb]
\subfloat

\bc
\bp
\psfrag{=}[Bc]{$=$}
\psfrag{*}[Bl]{$*$}
\psfrag{X}[Bc]{$x$}
\psfrag{sum}[Cl]{ $\sum$}
\psfrag{fi}[Bl]{$f_i^*$}
\psfrag{wi}[Bl]{$w_i^*$}
\psfrag{circf}[Bc]{$\quad \mathcal{F}^*$}
\psfrag{w}[Bl]{$w^*$}
\psfrag{(a)}[Bl]{\small{\bf(a)}Convolutional model}
\psfrag{(b)}[Bl]{\small{\bf(b)}Reformulated model}
\includegraphics[width=4in]{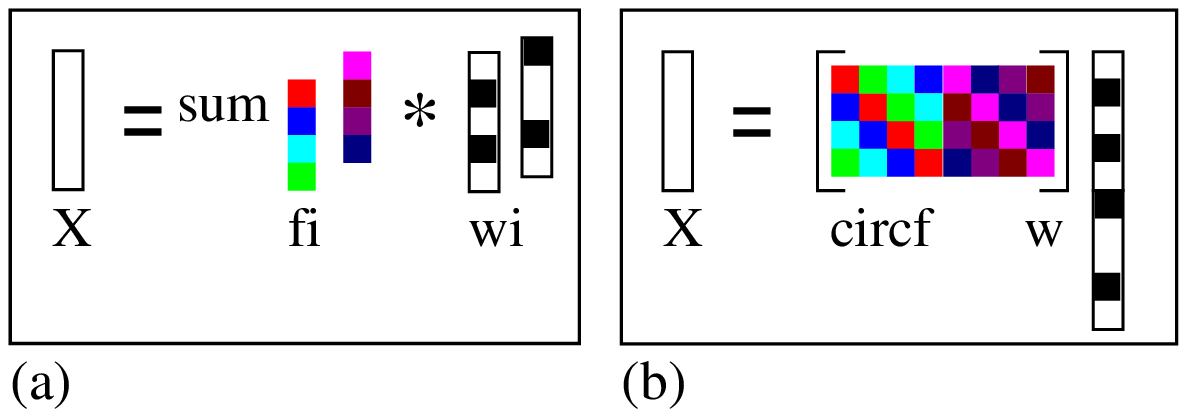}
\ep
\ec
\caption{Convolutional tensor decomposition for learning convolutional ICA models.(a) The convolutional generative model with 2 filters. (b) Reformulated model where $\CCir^*$ is column-stacked circulant matrix. }
\end{figure}

\begin{figure}
\bc
\bp
\psfrag{=}[Bc]{$=$}
\psfrag{+}[Bc]{$+ $}
\psfrag{..}[Bc]{{ $\ldots$}}
\psfrag{..}[Bc]{{ $\ldots$}}
\psfrag{....}[Bc]{{ $\ldots$}}
\psfrag{Tensor M3}[l]{$\Cum$}
\psfrag{lambda1}[l]{$\lambda_1 (\mathcal{F}_1^*)^{\otimes 3}$}
\psfrag{lambda2}[l]{$\lambda_2 (\mathcal{F}_2^*)^{\otimes 3}$}
\psfrag{lambda3}[l]{ $ \  \ldots$}
\psfrag{lambda4}[l]{ }
\includegraphics[width=4in]{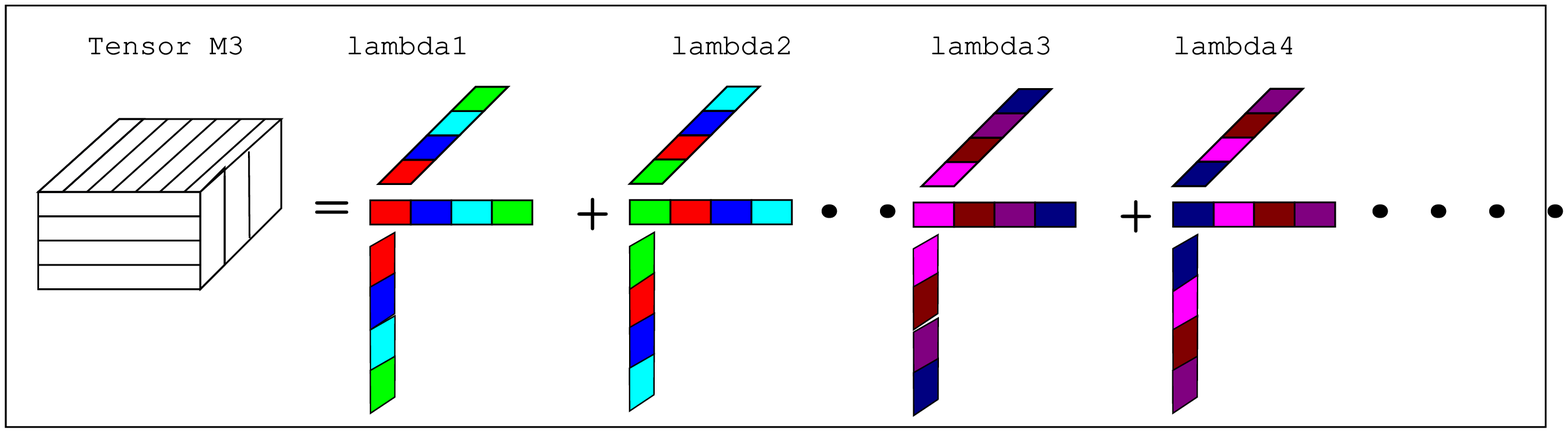}
\ep
\ec
\caption{Convolutional tensor decomposition for learning convolutional ICA models.  The third order cumulant is decomposed as filters. }
\end{figure}

We assume that the  input $x\in \mathbb{R}^n$ is generated as
\begin{equation}
x = \sum_{j\in [L]}   f_j^*\Conv w_j^*
	   = \sum_{j\in [L]} \Toep(f_j^*) w_j^*
	   = \CToep^*\cdot w^* ,\label{eqn:gen}
\end{equation}
where $\CToep^*:=[\Toep(f_1^*), \Toep(f_2^*),\ldots,\Toep(f_L^*)]$ is the concatenation or column stacked version  of circulant matrices and $w^*$ is the  \emph{row-stacked} vector $w^* : =[w_1^*;w_2^*;\ldots w_L^*]\in \Rbb^{nL}$.  Recall that $\Toep(f_l^*)$ is circulant matrix corresponding to filter $f_l^*$, as given by \eqref{eq:circulant_fft}. Note that although $\mathcal{F}^*$ is a $n$ by $nL$  matrix, there are only $nL$ free parameters. We never explicitly form the estimates $\mathcal{F}$ of $\mathcal{F}^*$, but instead use filter estimates  $f_l$'s to characterize $\mathcal{F}$.
In addition, we can handle additive Gaussian noise in \eqref{eqn:gen}, but do not incorporate it for simplicity.
\paragraph{Activation Maps: }For each observed sample $x$, the activation map $w_i^*$ in \eqref{eqn:gen} indicates the locations where each filter $f_i^*$ is active and $w^*$ is the  \emph{row-stacked} vector $w^* : =[w_1^*;w_2^*;\ldots w_L^*]$. We assume that the coordinates of $w^*$ are   drawn from some product distribution, i.e. different entries are independent of one another and we have the independent component analysis (ICA) model in \eqref{eqn:gen}. When the distribution encourages sparsity, e.g. Bernoulli-Gaussian, only a small subset of locations are active, and we have the {\em sparse coding} model in that case. We can also extend to dependent distributions such as Dirichlet for $w^*$, along the lines of~\cite{blei2003latent}, but limit ourselves to ICA model for simplicity.
\paragraph{Learning Problem: }Given access to $N$ i.i.d. samples,  $X:=[x^1,x^2,\ldots,x^N]\in \Rbb^{n\times N}$, generated according to the above model, we aim to estimate the true filters $f_i^*$, for $i \in [L]$. Once the filters are estimated, we can use standard decoding techniques, such as the square loss criterion in \eqref{eqn:alt-min} to learn the activation maps for the individual maps. We focus on developing novel methods for filter estimation in this paper.

\section{Form of  Cumulant Moment Tensors}\label{sec:CumForm}
\paragraph{Tensor Preliminaries} We consider 3rd order tensors in this paper but the analysis is easily extended to higher order tensors. For tensor $T\in\mathbb{R}^{n\times n\times n}$, its $(i_1,i_2,i_3)^\tha$ entry is denoted by $[T]_{i_1,i_2,i_3}, \forall i_1\in[n], i_2\in[n], i_3\in[n]$. 
%
A flattening or unfolding of tensor $T\in \Rbb$ is the column-stacked matrix of all its slices, given by $ \flatten(T):=[[T]_{:,:,1},[T]_{:,:,2},\ldots,[T]_{:,:,n}]\in\mathbb{R}^{n\times n^2}$. 
Define the Khatri-Rao product for vectors $u\in\mathbb{R}^{a}$ and $v\in\mathbb{R}^b$ as a \emph{row-stacked} vector $[u\odot v] := [u(1)v; u(2) v; \ldots; u(a)v] \in \mathbb{R}^{ab}$. Khatri-Rao product is also defined for matrices with same columns. For $M \in \mathbb{R}^{a\times c}$ and $M^\prime\in \mathbb{R}^{b\times c}$,  $M \odot M^\prime : = [M_1\odot M_1^\prime, \ldots, M_c\odot M_c^\prime, ]\in \mathbb{R}^{ab\times c}$, where $M_i$ denotes the $i^{\tha}$ column of $M$.


\paragraph{Cumulant: }The third order cumulant of a multivariate distribution   is a third order tensor, which uses (raw) moments up to third order. Let $\Cum\in \Rbb^{n \times n^2}$ denote the unfolded version of third order cumulant tensor, it is given by
\begin{equation}
\Cum : = \Ebb[x(x \odot x )^\top] - \flatten(Z) \label{eq:thirdCum_unfolding}
\end{equation}
where   $ [Z]_{a,b,c} := \Ebb[x_{a}]\Ebb[x_{b}x_{c}] +  \Ebb[x_{b}]\Ebb[x_{a}x_{c}] +  \Ebb[x_{c}]\Ebb[x_{a}x_{b}] -2 \Ebb[x_{a}]\Ebb[x_{b}]\Ebb[x_{c}], \ \forall a,b,c \in [n]. $

Under the convolution ICA model in Section~\ref{subsec:ConvolutionalICAmodel}, we show that the third order cumulant has a nice tensor form, as given below.

\begin{lemma}[Form of Cumulants]\label{lm:3orderMom}
The unfolded third order cumulant $\Cum$ in \eqref{eq:thirdCum_unfolding} has the following decomposition form
\begin{equation}\label{eqn:cumform}
\Cum = \sum_{j\in [nL]}\lambda_j^* \CToep^*_j (\CToep^*_j \odot\CToep^*_j)^\top={\mathcal{F}^*} \Lambda^* \left({\mathcal{F}^*}\odot {\mathcal{F}^*}\right)^\top, \quad \mbox{where }\Lambda^* := \Diag(\lambda_1^*,\lambda_2^*,\ldots,\lambda_{nL}^*)
\end{equation}
where $\CToep^*_j$ denotes the $j^{\tha}$ column of the
 \emph{column-stacked} circulant matrix  $\CToep^*$  and $\lambda^*_j$ is the third order cumulant corresponding to the (univariate) distribution of $w^*(j)$.
\end{lemma}

For example, if the $l^{\tha}$ activation is drawn  from a   Poisson distribution  with mean $\tilde{\lambda}$, we have that  $\lambda_l^*=\tilde{\lambda}$.
Note that if the third order cumulants  of the activations, i.e.  $\lambda_j^*$'s, are zero, we need to consider higher order cumulants. This holds for zero-mean activations and we need to use fourth order cumulant instead.  Our methods extend in a straightforward manner for higher order cumulants.

\begin{figure}[!htb]
\centering
\tikzstyle{matrx}=[rectangle,
                                    thick,
                                    minimum size=1.0cm,
                                    draw=gray!80,
                                    fill=gray!10]
   \tikzstyle{matrx_begin}=[rectangle,
                                    thick,
                                    minimum size=0.8cm]                                 
\tikzstyle{backgroun}=[rectangle,
                                                fill=gray!10,
                                                inner sep=0cm]

\begin{tikzpicture}[>=latex,text height=1.0ex,text depth=0.1ex]
  \matrix[row sep=0cm,column sep=0.0001cm] {
        \node(F)[matrx_begin]{{ $\CCir =$}};&
        \node (F1) [matrx] {{ $\block_1(\mathcal{F})$}};       &
        \node (F3)   [matrx] {{ $\quad \ldots\quad$}};       &
        \node (FL) [matrx] {{ $\block_L(\mathcal{F})$}};   \\    
            };
\end{tikzpicture}
\caption{{ Blocks of the column-stacked circulant matrix $ \CCir$.}}
\end{figure}
The decomposition form in \eqref{eqn:cumform} is known as the Candecomp/Parafac  (CP) decomposition form~\cite{anandkumar2014tensor} (the usual form has the decomposition of the tensor and not its unfolding, as above). We now attempt to recover the unknown filters $f_i^*$ through decomposition of the  third order cumulants $ \Cum$. This is formally stated below.

 
\paragraph{Objective Function: }
Our goal is to obtain filter estimates $f_i$'s which minimize   the Frobenius norm $\|\cdot\|_{\Fbb}$ of reconstruction of the   cumulant tensor $ \Cum$,
\begin{align}\nn&\min\limits_{\mathcal{F}}\quad \lVert \Cum - {\mathcal{F}} \Lambda \left({\mathcal{F}}\odot {\mathcal{F}}\right)^\top   \rVert^2_{\Fbb},
 \\ &\mbox{s.t. }   \block_l(\mathcal{F})
= U \Diag(\DFT(f_{l})) U\hermconj, \ \lVert f_l \rVert_2=1,\quad  \forall  l\in[L], \quad \Lambda=\Diag(\lambda).\label{eq:condition}
\end{align}
where $\block_l(\mathcal{F})$ denotes the $l^{\tha}$ circulant matrix   in $\CCir$. The conditions in \eqref{eq:condition} enforce $\block_l(\mathcal{F})$ to be circulant  and for the filters to be normalized. Recall that   $U$ denotes   the eigenvectors for circulant matrices. The rest of the paper is devoted to devising efficient methods to solve \eqref{eq:condition}.

Throughout the paper, we will use $\mathcal{F}_j$ to denote the $j^{\tha}$ column of $\mathcal{F}$, and $\block_l(\mathcal{F})$ to denote the $l^{\tha}$ circulant matrix block in $\mathcal{F}$. 
 Note that $\mathcal{F}\in \Rbb^{n\times nL}$, $\mathcal{F}_{j}\in \Rbb^{n}$ and $\block_l(\mathcal{F})\in\Rbb^{n\times n}$.


\section{Alternating Least Squares for Convolutional Tensor Decomposition}
To solve the non-convex optimization problem in \eqref{eq:condition}, we consider the alternating least squares (ALS) method with \emph{column stacked} circulant constraint. We first consider  the asymmetric relaxation of \eqref{eq:condition} and introduce separate variables $\modeA, \modeB$ and $\modeC$ for  filter estimates   along each of the modes to fit the third order cumulant  tensor $\Cum$.   We then perform alternating updates by fixing two of the modes and updating the third one.  For instance,
\begin{equation}\label{eqn:modeAopt}
\min\limits_{ {\modeA}} \quad  \lVert \Cum-  {\modeA} \Lambda \left({\modeC}\odot {\modeB}\right)^\top  \rVert^2_{\Fbb}\,\,
 \mbox{s.t. }   \block_l(\modeA) 
= U \cdot\Diag(\DFT(f_{l}))\cdot U\hermconj, \ \lVert f_{l} \rVert_2^2=1,  \forall  l\in[L]
\end{equation} Similarly, $\modeB$ and $\modeC$ have the same column-stacked circulant matrix constraint and  are updated similarly in alternating steps. The diagonal matrix $\Lambda$ is updated through normalization.
%

We now introduce the \emph{Convolutional Tensor } ($\ouralgorithm$) Decomposition algorithm to efficiently solve~\eqref{eqn:modeAopt} in closed form, using simple operations such as matrix multiplications and fast Fourier Transform (FFT). We do not form matrices $\modeA, \modeB$ and $\modeC\in \Rbb^{n \times nL}$, which are large, but only update them using filter estimates $f_1, \ldots, f_L, g_1, \ldots, g_L, h_1, \ldots h_L$.

Using the property of least squares,  the optimization problem in \eqref{eqn:modeAopt}
is equivalent to \begin{equation}\label{eqn:ls}
\min\limits_{ {\modeA}}    \lVert \Cum (\left({\modeC}\odot {\modeB}\right)^\top)^\dag \Lambda^\dag-  {\modeA}    \rVert^2_{\Fbb}\,\,
 \mbox{s.t. }   \block_l(\modeA) 
= U \cdot\Diag(\DFT(f_{l}))\cdot U\hermconj, \ \lVert f_{l} \rVert_2^2=1,  \forall  l\in[L]\eeq
when  $(\modeC\odot \modeB)$ and $\Lambda$ are full column rank.
The full rank condition requires $nL<n^2$ or $L <n$,  and it is a reasonable assumption since otherwise the filter estimates are redundant. In practice, we can additionally regularize the update to ensure full rank condition is met. Denote
\begin{equation}\label{eqn:newCum}
\newCum : =  \Cum (({\modeC}\odot {\modeB})^\top)^\dag, 
\end{equation}  where $\dag$ denotes 
pseudoinverse. Let $\block_l(\newCum)$ and $\block_l(\Lambda)$   denote the $l^{\tha}$  blocks of $\newCum$ and $\Lambda$.
Since \eqref{eqn:ls} has block constraints, it can be broken down in to solving $L$ independent sub-problems
\begin{equation}\label{eqn:newCumopt}
  \min_{f_{l}}
\left\lVert
 \block_l(\newCum)\cdot \block_l(\Lambda)^\dag - U \cdot\Diag(\DFT(f_l))\cdot U\hermconj
  \right\rVert^2_{\Fbb} \\
\quad  s.t. \quad \lVert f_l \rVert_2^2=1, \forall l\in[L]
\end{equation}

We present the main result now.
\begin{theorem}[Closed form updates]The optimal solution $f_l^{\opt}$ for \eqref{eqn:newCumopt} is given by
\begin{align}\label{eq:circulantProjection}
f_l^{\opt} (p) &= \frac{\sum\limits_{i,j\in[n]} \| \block_l(\newCum)_{j}\|^{-1} \cdot  \block_l(\newCum)_{j}^i\cdot I_{p-1}^{q }}{\sum\limits_{i,j\in [n]}   I_{p-1}^{q} }, &\forall p\in[n], q:=(i-j)\mod n.
\end{align} Further $\Lambda=\Diag(\lambda)$ is updated as
$\lambda(i) = \|\newCum_i\|$, for all  $i\in[nL]$.
\end{theorem} 

Thus, the reformulated problem in \eqref{eqn:newCumopt} can be solved in closed form efficiently. A bulk of the computational effort will go into computing $\newCum$ in \eqref{eqn:newCum}.
Computation of $ \newCum$ requires $2L$ fast Fourier Transforms of length $n$ filters and simple matrix multiplications without explicitly forming ${\modeB}$ or ${\modeC}$. We make this concrete in the next section. The closed form update after getting $\newCum$ is highly parallel. With $O(n^2L/\log n)$ processors, it takes $O(\log n)$ time. 
\section{Algorithm Optimization for Reducing Memory and Computational Costs}
We now focus on estimating    $\,\newCum:=  \Cum (({\modeC}\odot {\modeB})^\top)^\dag$  in \eqref{eqn:newCum}. If done naively, this requires inverting $n^2 \times nL$ matrix and multiplication of $n \times n^2$ and $n^2 \times nL$ matrices with $O(n^6)$ time. However, forming and computing with these matrices is very expensive when $n$ (and $L$) are large. Instead, we utilize the properties of circulant matrices and the Khatri-Rao product $\odot$ to efficiently carry out these computations implicitly. We present our final result on computational complexity of the proposed method.
\begin{lemma}[\textbf{Computational Complexity}]\label{lm:computationalComplexity}
With multi-threading,  the running time is $O(\log n +\log L)$ per iteration using $O(L^2n^3)$ processes.
\end{lemma}
We now describe how we utilize various algebraic structures to obtain efficient computation.

\paragraph{Property 1 }(Khatri-Rao product): $(({\modeC}\odot {\modeB})^\top)^\dag = ({\modeC}\odot{\modeB}) (({\modeC}^\top{\modeC}) .\star ({\modeB}^\top {\modeB}) )^\dag $, where $.\star$  denotes element-wise product.

\paragraph{Computational Goals:  }Find   $(({\modeC}^\top{\modeC}) .\star ({\modeB}^\top {\modeB}) )^\dag$ first and multiply the result with $ \Cum({\modeC}\odot{\modeB})$ to find $\newCum$.

\smallskip
We now describe in detail how to carry out each of these steps.
\subsection{Challenge: Computing $(({\modeC}^\top{\modeC}) .\star ({\modeB}^\top {\modeB}) )^\dag$}

A naive implementation to find the matrix inversion  $(({\modeC}^\top{\modeC}) .\star ({\modeB}^\top {\modeB}))^\dag$ is 
very expensive. However, we incorporate the stacked circulant structure of $\modeB$ and $\modeC$ to reduce computation. Note that this is not completely straightforward since although $\modeB$ and $\modeC$ are column stacked circulant matrices, the resulting product whose inverse is required, is {\em not} circulant. Below, we show that however, it is partially circulant along different rows and columns.

\paragraph{ Property 2 } (Block circulant matrix): The matrix $({\modeC}^\top{\modeC}) .\star ({\modeB}^\top {\modeB})$ consists of row and column stacked circulant matrices.

We now make the above property precise by introducing some new notation. Define column stacked identity matrix $\mathbf{I} : = [I,\ldots,I]\in \mathbb{R}^{n\times nL}$, where $I$ is $n \times n $ identity matrix. Let $\bfU:=\blkdiag(U, U, \ldots U) \in \Rbb^{n L \times nL}$ be the block diagonal matrix with $U$ along the diagonal. The first thing to note is that $\modeB$ and $\modeC$, which are column stacked circulant matrices, can be written as \beq\label{eqn:conc} \modeB = \bfI \cdot \bfU \cdot \Diag(v) \cdot\bfU^{\hermconj}, \quad v:=[\DFT(g_1); \DFT(g_2);\ldots; \DFT(g_L)], \eeq where $g_1$, \ldots $g_L$ are the filters corresponding to $\modeB$, and similarly for $\modeC$, where the diagonal matrix consists of FFT coefficients of the respective filters $h_1, \ldots, h_L$.

\begin{figure}[!htb]
\centering
\tikzstyle{matrx}=[rectangle,
                                    thick,
                                    minimum size=0.8cm,
                                    draw=gray!80,
                                    fill=gray!10]
   \tikzstyle{matrx_begin}=[rectangle,
                                    thick,
                                    minimum size=0.8cm]                                 
\tikzstyle{backgroun}=[rectangle,
                                                fill=gray!10,
                                                inner sep=0cm]
\begin{tikzpicture}[>=latex]
  \matrix[row sep=0.0000cm,column sep=0.000cm] {
        & \node (F11) [matrx] {{$\ \block_1^1(\mathbf{\Psi})$}};       &
        \node (F12)   [matrx] {{$\quad\  \ldots\quad$}};        &
        \node (F13) [matrx] {{$\block_L^1(\mathbf{\Psi})$}};       
        \\
        \node(F)[matrx_begin]{{$\mathbf{\Psi} =$}}; & 
         \node (F21)   [matrx] {{$ \ \quad\  \ldots\quad$}};       &
          \node (F22)   [matrx] {{$\quad  \ \ldots\quad$}};       &
           \node (F23)   [matrx] {{$\ \quad  \ldots\ \quad$}};        &
           \\
           & \node (F31) [matrx] {{$\block_1^L(\mathbf{\Psi})$}};       &
        \node (F32)   [matrx] {{$\quad\  \ldots\quad$}};       &
        \node (F33) [matrx] {{$\block_L^L(\mathbf{\Psi})$}};   
        \\
};
\end{tikzpicture}
\caption{{ Blocks of the row-and-column-stacked diagonal matrices $ \mathbf{\Psi}$. $\block_j^i(\mathbf{\Psi})$ is diagonal.}}
\end{figure}
By appealing to the above form, we have the following result. We use the notation $\block_j^i(\mathbf{\Psi})$ for a matrix $\mathbf{\Psi}\in \Rbb^{nL\times nL}$ to denote $(i,j)^{\tha}$ block of size $n \times n$.
\begin{lemma}[Form of $({\modeC}^\top{\modeC}) .\star ({\modeB}^\top {\modeB}) $ ]We have
	 \begin{equation}\label{eq:rowcolumnstackdecomp}
	 (({\modeC}^\top{\modeC}) .\star ({\modeB}^\top {\modeB})  )^\dag = \mathbf{U} \cdot\mathbf{\Psi}^\dag\cdot \mathbf{U}\hermconj,
	 \end{equation}
	 where $\mathbf{\Psi}\in \Rbb^{nL \times nL}$ has $L$ by $L$ blocks, each block of size $n \times n$. Its $(j,l)^{\tha}$  block is given by
	 	\begin{equation}
	 	\block_{l}^j(\mathbf{\Psi})  = \Diag\hermconj(\DFT(g_{j}))\cdot\Diag\hermconj(\DFT(h_j))\cdot\Diag(\DFT(g_l))\cdot\Diag(\DFT(h_l))\in \Rbb^{n \times n}
	 	\end{equation}
\end{lemma}

Therefore, the inversion of $({\modeC}^\top{\modeC}) .\star ({\modeB}^\top {\modeB}) $ can be reduced to the inversion of row-and-column stacked set of  diagonal matrices which form $\mathbf{\Psi}$. Computing $\mathbf{\Psi}$  simply requires  FFT on all $2L$ filters $g_1, \ldots, g_L$ and $h_1,\ldots, h_L$, i.e. $2L$ FFTs, each on length $n$ vector.
We propose an efficient iterative algorithm to compute $\mathbf{\Psi}^\dag$ via block matrix inversion theorem\cite{golub2012matrix} in Appendix~\ref{appdx:parallelBlockInversion}.

\subsection{Challenge: Computing $\newCum= \Cum({\modeC}\odot{\modeB})\cdot(({\modeC}^\top{\modeC}) .\star ({\modeB}^\top {\modeB}) )^\dag$}

 Now that we have computed $(({\modeC}^\top{\modeC}) .\star ({\modeB}^\top {\modeB})  )^\dag$ efficiently, we need to compute the resulting matrix with $ \Cum({\modeC}\odot{\modeB})$ to obtain $\newCum$. 
We observe that the $m^{\tha}$ row of the result $\newCum$ is given by
\begin{align}\label{eqn:M}
M^m= \sum_{j\in [nL]} {\mathbf{U}}^j \Diag\hermconj\left(z\right) \Phi^{(m)} \Diag \left(v\right) ({\mathbf{U}}^j)\hermconj {\mathbf{U}}^j  \mathbf{\Psi}^\dag {\mathbf{U}}\hermconj, \quad \forall m\in [nL],
\end{align}
where $v:=[\DFT(g_1);\ldots; \DFT(g_L)]$, $z:=[\DFT(h_1);\ldots; \DFT(h_L)]$ are concatenated FFT coefficients of the filters, and
\begin{align} \Phi^{(m)}&:={\mathbf{U}}\hermconj \mathbf{I}^\top  \Gamma^{(m)} \mathbf{I} {\mathbf{U}}, \quad 
[\Gamma^{(m)}]_j^i:= [\Cum]_{i+(j-1)n}^m, \quad \forall i,j,m\in[n] \end{align}
Note that $\Phi^{(m)}$ and $\Gamma^{(m)}$ are fixed for all iterations and needs to be computed only once. Note that $\Gamma^{(m)}$ is the result of taking $m^{\tha}$ row of the cumulant unfolding $\Cum$ and matricizing it. Equation~\eqref{eqn:M} uses the property that $\Cum^m(\modeC\odot \modeB)$ is equal to the diagonal elments of $ \modeC^\top \Gamma^{(m)} \modeB$.

We now bound the cost for computing \eqref{eqn:M}. (1) Inverting $ \mathbf{\Psi}$  takes $O(\log L +\log n)$ time with $O(n^2L^2/(\log n + \log L))$ processors 
according to appendix~\ref{appdx:parallelBlockInversion}.
(2) Since $\Diag(v)$ and $\Diag(z)$ are diagonal and $\mathbf{\Psi}$ is a matrix with diagonal blocks, the overall matrix multiplication in equation~\eqref{eqn:M} takes $O(L^2n^2)$ time serially with $O(L^2n^2)$ degree of parallelism for each row. Therefore the overall serial computation cost is $O(L^2n^3)$ with $O(L^2n^3)$ degree of parallelism.  With multi-threading, the running time is $O(1)$ per iteration using $O(L^2n^3)$ processes. (3) $\DFT$ requires $O(n\log n)$ serial time, with $O(n)$ degree of parallelism. Therefore computing $2L$ $\DFT$'s takes $O(\log n)$ time with $O(Ln)$ processors. 

Combining the above discussion, it takes $O(\log L+\log n)$ time with $O(L^2n^3)$ processors.

\section{Experiments: Compare with Alternating Minimization}
\begin{figure}[!htb]
\bc 
\psfrag{error}{\small{error}}
\psfrag{filter 1 2 reconstruction error}{}
\psfrag{iteration}{\small{iteration}}
\psfrag{tensor method reconstruction error}[Bl]{\small{Proposed $\mathsf{CT}$: Reconst}}
\psfrag{alternating minimization reconstruction error}[Bl]{\small{Baseline $\mathsf{AM}$: Reconst}}
\psfrag{tensor method for filter 1}[Bl]{\small{Proposed $\mathsf{CT}$: $f_1$}}
\psfrag{alternating minimization for filter 1}[Bl]{\small{Baseline $\mathsf{AM}$: $f_1$}}
\psfrag{tensor method for filter 2}[Bl]{\small{Proposed $\mathsf{CT}$: $f_2$}}
\psfrag{alternating minimization for filter 2}[Bl]{\small{Baseline $\mathsf{AM}$: $f_2$}}
\includegraphics[width=4in]{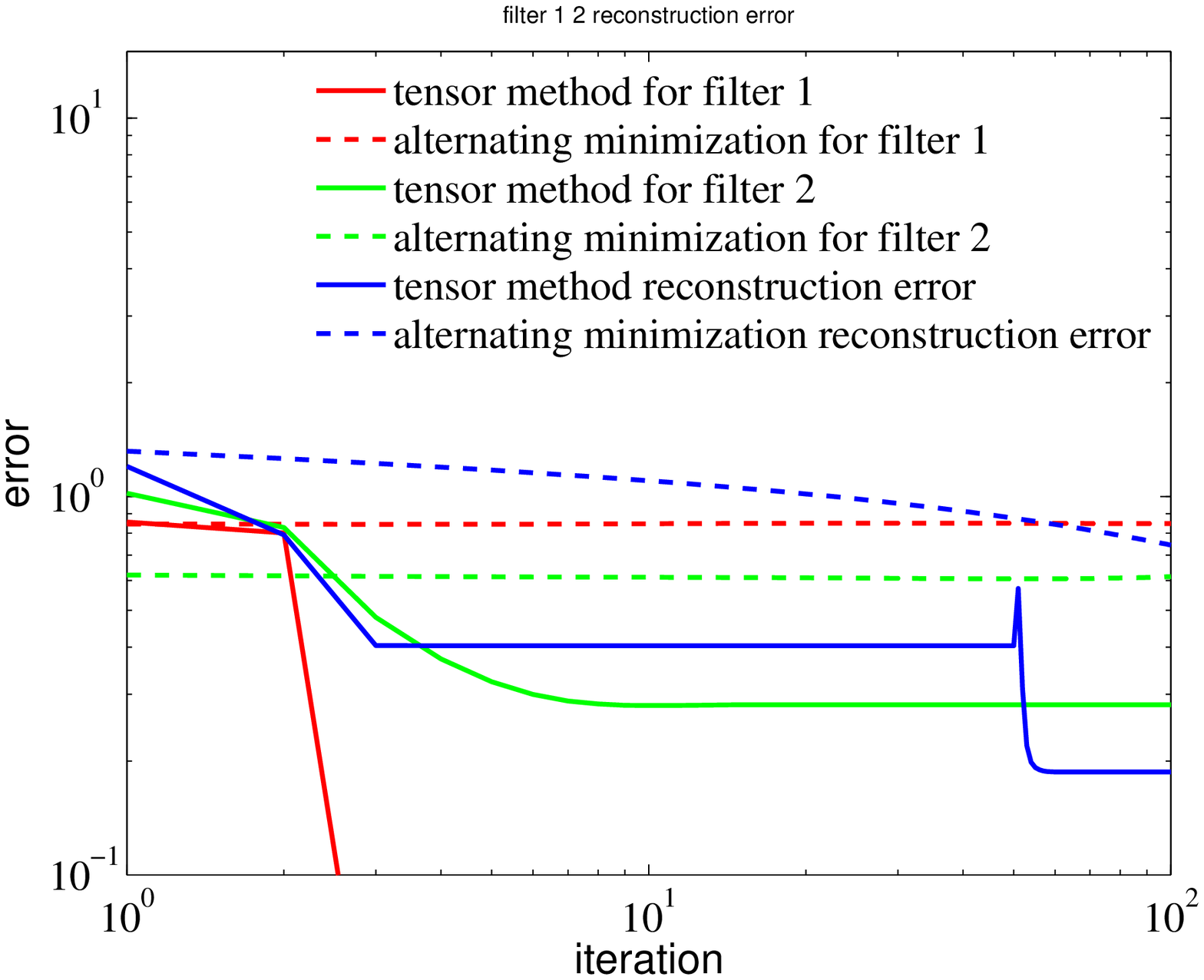}\ec
\caption{Error comparison (on filters and reconstruct tensor) between our convolutional tensor method (proposed $\mathsf{CT}$) and the baseline alternate minimization method (baseline $\mathsf{AM}$).}
\label{fig:error}
\end{figure}

\begin{figure}[!htb]
\subfloat[b][Running Times Scale with $L$]
{\begin{minipage}{3in}\bc 
\psfrag{Number of Filters L}[Bl]{\small{Number of Filters $L$}}
\psfrag{running time in seconds}[Bl]{\small{seconds}}
\psfrag{tensor method}[Bl]{\small{Proposed $\mathsf{CT}$}}
\psfrag{alternating minimization}[Bl]{\small{Baseline $\mathsf{AM}$}}
\includegraphics[width=3in]{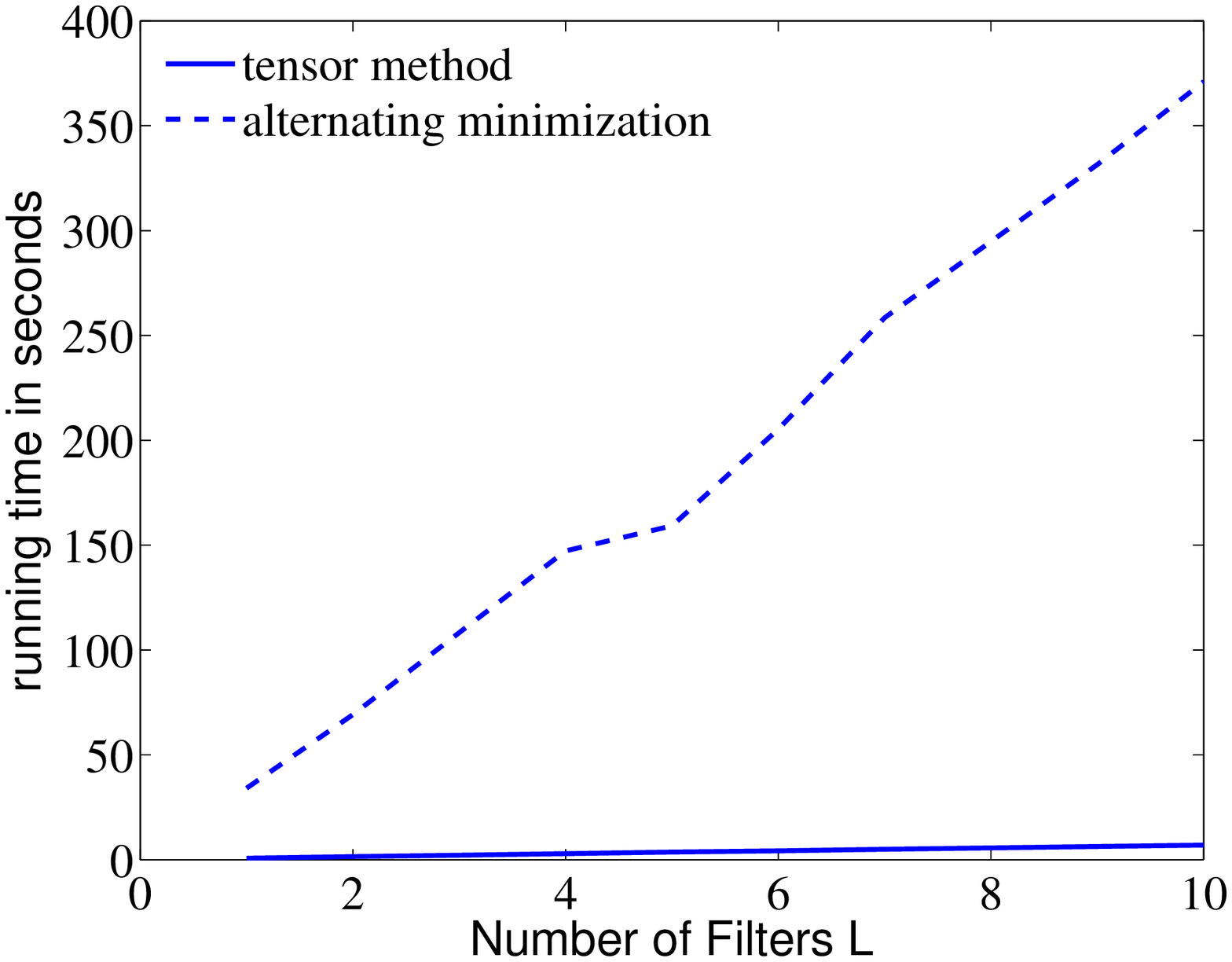}\ec\end{minipage}}\hfil
\subfloat[c][Running Times Scale with $N$]
{\begin{minipage}{3in}\bc  
\psfrag{Number of Samples N}[Bl]{\small{Number of Samples $N$}}
\psfrag{running time in seconds}[Bl]{\small{seconds}}
\psfrag{tensor method}[Bl]{\small{Proposed $\mathsf{CT}$}}
\psfrag{alternating minimization}[Bl]{\small{Baseline $\mathsf{AM}$}}
\includegraphics[width=3in]{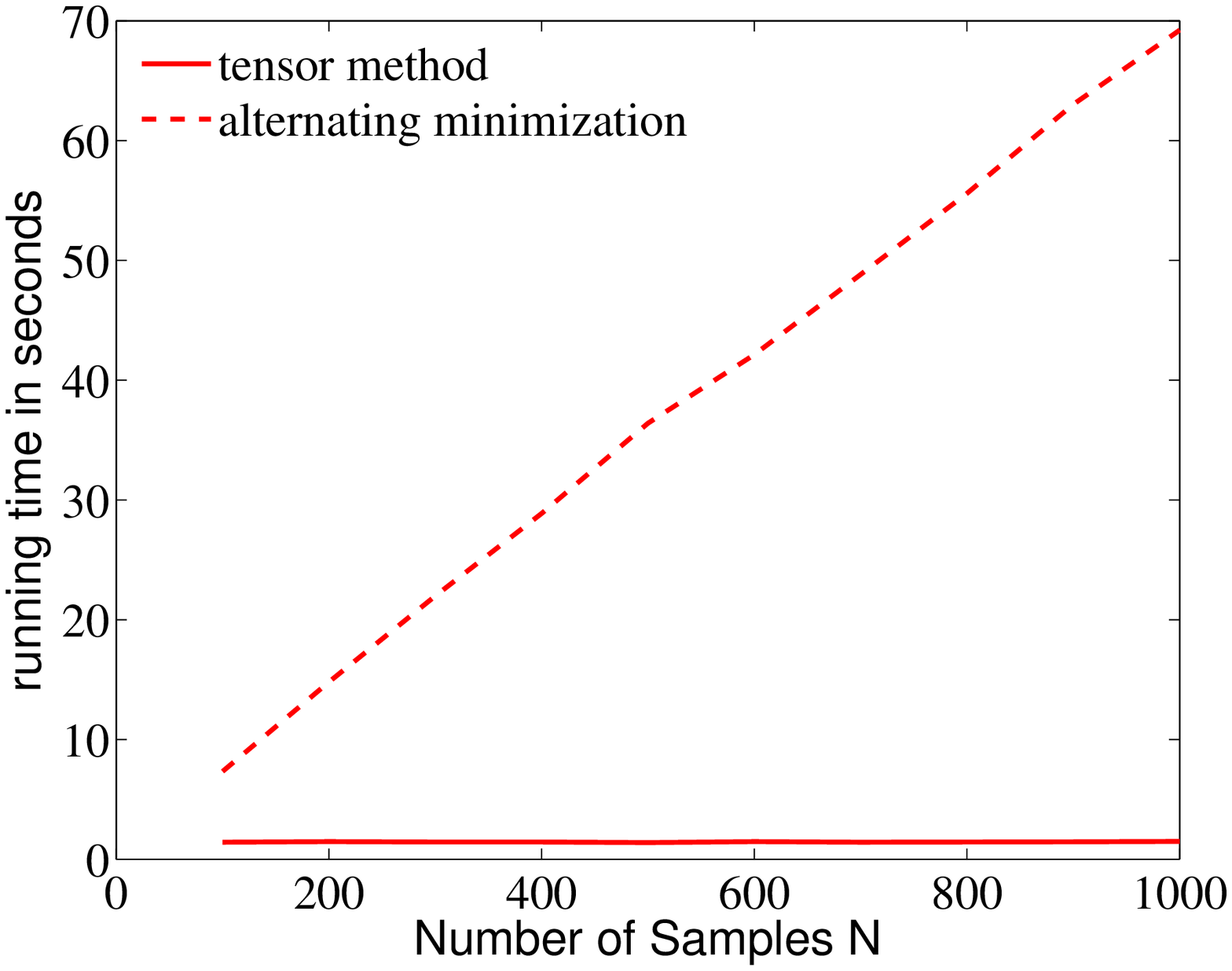}\ec\end{minipage}}\\
		\caption{{ (a) Running time comparison between our proposed $\mathsf{CT}$ method and the baseline $\mathsf{AM}$ method under varying $L$. (b) Running time comparison between our proposed $\mathsf{CT}$ method and the baseline $\mathsf{AM}$ method under varying $N$.}}
\label{fig:runtime}
\end{figure}	
We compare our convolutional tensor decomposition framework with alternating (between filters and activation map) minimization method in equation~\eqref{eqn:alt-min} where
gradient descent is employed to update $f_i$ and $w_i$ alternatively. 

The error comparison between our proposed convolutional tensor algorithm and the alternating minimization algorithm is in figure~\ref{fig:error}. We evaluate the errors for both algorithms by comparing the reconstruction of error and filter recovery error. Our algorithm converges much faster to the solution than the alternating minimization algorithm. In fact, the alternating minimization  leads to spurious solution where the reconstruction error decreases but filter estimation error increases. Note that in experiments, we observe a more robust performance if we do deflation, meaning that we recover the filters one by one. The error bump in the reconstruction error curve in figure~\ref{fig:error} is due to the random initialization. 

The running time is also reported in figure~\ref{fig:runtime} between our proposed convolutional tensor algorithm and the alternating minimization. Our algorithm is orders of magnitude faster than the alternating minimization. Both our algorithm and alternating minimization scale linearly with number of filters. However convolutional tensor algorithm scales constantly with the number of samples whereas the alternating minimization scales linearly.

\section{Discussion}

There are number of questions for future investigation. 
We plan to extend learning sentence embeddings to handle more challenging tasks involving long documents such as plagiarism detection and reading comprehension since our framework is scalable.  Since our framework is parallelizable, it can handle long vectors and learn in an efficient manner.
Our framework easily extends to convolutional models for higher dimensional signals, where the circulant matrix is replaced with block circulant matrices~\cite{gray2005toeplitz}. More generally, our framework can in principle handle general group structure which are diagonalizable. Deploying the frameworks in settings which incorporate Lie algebra is of great practical interest in computer vision and robotics.
By combining the advantages of tensor methods with a general class of invariant representations, we expect to have a powerful paradigm for learning efficient embeddings.


\subsubsection*{Acknowledgments}
We thank Cris Cecka for helpful discussion on fast implementation of block matrix inverse and initial discussions with Majid Janzamin and Hanie Sedghi on Toeplitz matrices.
{
\bibliographystyle{plain}\bibliography{0arxiv_tensorconv}
}
\normalsize
\newpage
\begin{center}
\textbf{\LARGE Appendix for Convolutional Dictionary Learning through Tensor Factorization}
\end{center}
\appendix
\section{Cumulant Form}\label{appdx:momentForm}
 In~\cite{anandkumar2014tensor}, it is proved that in ICA model, the cumulant of observation $x$ is decomposed into multi-linear transform of a diagonal cumulant of $h$. Therefore, we aim to find the third order cumulant for input $x$. 
 
  As we know that the $r^\tha$ order moments for variable $x$ is defined as 
\begin{equation}
\mu_r : = \mathbb{E}[x^r]\in \mathbb{R}^{n\times n\times n}
\end{equation} 
Let us use $[\mu_3]_{i,j,k}$ to denote the $(i,j,k)^\tha$ entry of the third order moment. The relationship between $3^\tha$ order cumulant $\kappa_3$and $3^\tha$ order moment $\mu_3$is 
\begin{align}
[\kappa_3]_{i,j,k} = [\mu_3]_{i,j,k} -  [\mu_2]_{i,j}[\mu_1]_{k} -  [\mu_2]_{i,k}[\mu_1]_{j} -  [\mu_2]_{j,k}[\mu_1]_{i}+ 2 [\mu_1]_{i}[\mu_1]_{j}[\mu_1]_{k}
\end{align}

Therefore the shift tensor is in this format: 
We know that the shift term
\begin{align}
[Z]_{a,b,c} := \Ebb[x^i_{a}]\Ebb[x^i_{b}x^i_{c}] +  \Ebb[x_{b}]\Ebb[x_{a}x^i_{c}] +  \Ebb[x_{c}]\Ebb[x_{a}x_{b}] -2 \Ebb[x_{a}]\Ebb[x_{b}]\Ebb[x_{c}], \quad a,b,c \in [n]
 \end{align}

It is known from~\cite{anandkumar2014tensor} that cumulant decomposition in the 3 order tensor format is 
\begin{equation}
\Ebb[x\otimes x \otimes x] - Z = \sum_{j\in[nL]}\lambda_j^* \CCir_j^* \otimes \CCir_j^* \otimes \CCir_j^*
\end{equation}

Therefore using the Khatri-Rao product property,
\begin{equation}
\flatten(\sum_{j\in[nL]}\lambda_j^* \CCir_j^* \otimes \CCir_j^* \otimes \CCir_j^*) = \sum_{j\in [nL]}\lambda_j^* \CToep^*_j (\CToep^*_j \odot\CToep^*_j)^\top ={\mathcal{F}^*} \Lambda^* \left({\mathcal{F}^*}\odot {\mathcal{F}^*}\right)^\top
\end{equation}

Therefore the unfolded third order cumulant is decomposed as 
$\Cum ={\mathcal{F}^*} \Lambda^* \left({\mathcal{F}^*}\odot {\mathcal{F}^*}\right)^\top$.

\section{Parallel Inversion of $\mathbf{\Psi}$}
\label{appdx:parallelBlockInversion}
We propose an efficient iterative algorithm to compute $\mathbf{\Psi}^\dag$ via block matrix inversion theorem\cite{golub2012matrix}.

\begin{lemma}(Parallel Inversion of row and column stacked diagonal matrix)\label{lemma:matrix_inverse}
	Let $ J^L = \mathbf{\Psi} $ be partitioned into a block form:
	\begin{equation}
	J^L = \left[
	\begin{array}{cc}
	J^{L-1} & O \\
	R & \block_L^L(\mathbf{\Psi}) \\
	\end{array}
	\right],
	\end{equation}
	where  $O : =  \left[
	\begin{array}{c}
	\block_L^1(\mathbf{\Psi}) \\
	 \vdots\\
	 \block_L^{L-1}(\mathbf{\Psi})
	\end{array}
	\right]$, and $R : = \left[ \block_{L-1}^1(\mathbf{\Psi}),\ldots,\block_{L-1}^{L}(\mathbf{\Psi})\right]$. After inverting $\block_L^L(\mathbf{\Psi})$ which takes $O(1)$ time using $O(n)$ processors, there inverse of  $\mathbf{\Psi}$ is achieved by 
	\begin{equation}\label{eq:recursiveInverseBlock}
	\mathbf{\Psi}^\dag = \left[
	\begin{array}{ll}
	(J^{L-1}-O{\block_L^L(\mathbf{\Psi})}^{-1}R)^{-1} & -{(J^{L-1})}^{-1}O({\block_L^L(\mathbf{\Psi})}-R{(J^{L-1})}^{-1}O)^{-1} \\
	-{\block_L^L(\mathbf{\Psi})}^{-1}R(J^{L-1}-O{\block_L^L(\mathbf{\Psi})}^{-1}R)^{-1} & ({\block_L^L(\mathbf{\Psi})}-R{(J^{L-1})}^{-1}O)^{-1}
	\end{array}
	\right]
	\end{equation}
	 assuming that $J^{L-1}$ and $\block_L^{L}\mathbf{\Psi}$ are invertible. 
	 \end{lemma}
	 This again requires inverting $R$, $O$ and $J^{L-1}$. Recursively applying these block matrix inversion theorem, the inversion problem is reduced to  inverting $L^2$ number of $n$ by $n$ diagonal matrices with additional matrix multiplications as indicated in equation~\eqref{eq:recursiveInverseBlock}.

Inverting a diagonal matrix results in another diagonal one, and the complexity of inverting $n\times n$ diagonal matrix is $O(1)$ with $O(n)$ processors. We can simultaneous invert all blocks. Therefore with $O(nL^2)$ processors, we invert all the diagonal matrices in $O(1)$ time. The recursion takes $L$ steps, for step $i \in[L]$ matrix multiplication  cost is O($\log nL$) with $O(n^2 L/\log (nL))$ processors. With $L$ iteration, one achieves $O(\log n + \log L)$ running time with $O(n^2L^2/(\log L + \log n))$ processors.

\end{document}